\documentclass{article}
\usepackage{iclr2026_conference,times}

\usepackage{amsmath,amsfonts,bm}

\def\eqref#1{equation~\ref{#1}}

\def\1{\bm{1}}

\DeclareMathAlphabet{\mathsfit}{\encodingdefault}{\sfdefault}{m}{sl}
\SetMathAlphabet{\mathsfit}{bold}{\encodingdefault}{\sfdefault}{bx}{n}

\usepackage{hyperref}
\usepackage{url}
\usepackage{booktabs}
\usepackage{amsfonts}
\usepackage{nicefrac}
\usepackage{microtype}
\usepackage[table, svgnames]{xcolor}
\usepackage{amsmath}
\usepackage{amssymb}
\usepackage{pgfplots}
\usepackage{pgf-pie}
\usepackage{siunitx}
\usepgfplotslibrary{fillbetween}
\usetikzlibrary{pgfplots.groupplots}
\pgfplotsset{compat=1.9}
\usepackage[titlenumbered,ruled,linesnumbered]{algorithm2e}
\let\oldnl\nl
\newcommand{\nonl}{\renewcommand{\nl}{\let\nl\oldnl}}
\usepackage{graphicx}
\graphicspath{ {images/} }
\usepackage{multirow}
\usepackage{makecell}
\usepackage{enumitem}
\usepackage{soul}
\usepackage[export]{adjustbox}
\usepackage{bbm}
\usepackage{bbding}
\usepackage{minitoc}
\usepackage{subfig}
\usepackage{cleveref}
\usepackage{mathtools}
\usepackage{tcolorbox}
\usepackage{amsthm}
\usepackage{enumitem}

\title{Digging Deeper: Learning Multi-Level Concept Hierarchies}

\author{Oscar Hill\textsuperscript{\textdagger}, Mateo Espinosa Zarlenga\textsuperscript{\textdagger,\textasteriskcentered}, Mateja Jamnik\textsuperscript{\textdagger}\\
\textsuperscript{\textdagger}Department of Computer Science and Technology, University of Cambridge, UK\\
\textsuperscript{\textasteriskcentered}Department of Computer Science, University of Oxford, UK\\
\texttt{\scriptsize ogh22@cam.ac.uk, mateo.espinosazarlenga@trinity.ox.ac.uk, mateja.jamnik@cl.cam.ac.uk}}

\iclrfinalcopy
\begin{document}

\maketitle

\begin{abstract}

Although concept-based models promise interpretability by explaining predictions with human-understandable concepts, they typically rely on exhaustive annotations and treat concepts as flat and independent. To circumvent this, recent work has introduced \textit{Hierarchical Concept Embedding Models} (HiCEMs) to explicitly model concept relationships, and \textit{Concept Splitting} to discover sub-concepts using only coarse annotations. However, both HiCEMs and Concept Splitting are restricted to shallow hierarchies. We overcome this limitation with \textit{Multi-Level Concept Splitting} (MLCS), which discovers multi-level concept hierarchies from only top-level supervision, and \textit{Deep-HiCEMs}, an architecture that represents these discovered hierarchies and enables interventions at multiple levels of abstraction. Experiments across multiple datasets show that MLCS discovers human-interpretable concepts absent during training and that Deep-HiCEMs maintain high accuracy while supporting test-time concept interventions that can improve task performance.

\end{abstract}

\section{Introduction}

Concept-based models \citep{cbm, cem, labelfree} aim to make neural networks interpretable by predicting human-understandable concepts and using them to explain decisions. By exposing intermediate concepts, these models allow users to inspect, debug, and intervene on model reasoning. However, most concept-based approaches assume concepts are flat and independent, leading to their representations failing to capture known inter-concept relationships \citep{naveen}. This is problematic because real-world concepts are often interrelated, and human cognition utilises these relationships \citep{cognitive}.

To address this limitation, \textit{Hierarchical Concept Embedding Models} (HiCEMs, \citet{hicem}) explicitly model concept relationships and utilise a procedure called \textit{Concept Splitting} to reduce annotation costs by discovering sub-concepts using only coarse top-level concept labels. However, these methods are restricted to shallow hierarchies, supporting only a single layer of sub-concepts in addition to the provided coarse concepts. This prevents models from capturing richer structure and from offering interventions at several levels of abstraction.

To overcome these issues, we propose \textit{Multi-Level Concept Splitting} (MLCS), a method for discovering multi-level concept hierarchies from concept embeddings trained only with top-level supervision, and \textit{Deep-HiCEM}, an architecture designed to support arbitrarily deep concept hierarchies. Together, these contributions enable multi-level hierarchical explanations and interventions in concept-based models without requiring exhaustive concept annotations.

We evaluate our approach across several datasets, including a synthetic dataset designed for hierarchical concept discovery. Our results show that multi-level human-interpretable concept hierarchies can be discovered reliably, and used to explain predictions without sacrificing performance. Our contributions are:

\begin{itemize}
    \item We introduce \textbf{MLCS} for discovering multi-level concept hierarchies from only top-level supervision.
    \item We propose \textbf{Deep-HiCEMs}, which model arbitrarily deep concept hierarchies and allow human interventions at any level in the hierarchy.
    \item We demonstrate that Deep-HiCEMs trained via MLCS can accurately discover interpretable concept hierarchies that were absent during training. Moreover, our experiments show that Deep-HiCEMs trained with MLCS achieve competitive task accuracies and are receptive to test-time concept interventions at different levels of granularity.
\end{itemize}

\section{Background and Related Work}
\label{background}

\paragraph{Concept learning} Concept-based methods (CMs) aim to explain model predictions using human-understandable concepts (e.g., ``colour'' or ``size'') \citep{dissect, net2vec, tcav}. Some CMs, such as Concept Bottleneck Models (CBMs, \citet{cbm}) and Concept Embedding Models (CEMs, \citet{cem, mixcem}), explicitly incorporate concepts into model architectures, leading to inherently interpretable models that provide concept-based explanations. These models, however, typically require large concept-annotated training sets and may suffer from suboptimal predictive performance when concept labels are incomplete~\citep{cem} or noisy~\citep{penaloza2025addressing}. Furthermore, most CMs ignore relationships between concepts, instead assuming all concepts are independent~\citep{havasi2022addressing,vandenhirtz2024stochastic, causal_cgm}.

\paragraph{Concept Splitting and HiCEMs} Concept Splitting \citep{hicem} is a method for automatically discovering fine-grained sub-concepts from a model trained with only coarse concept labels. It uses sparse autoencoders \citep{monosemanticity} to discover sub-concepts in concept-aligned embeddings taken from a pretrained CEM \citep{cem}. This reveals interpretable, finer concepts without extra annotations, enabling more granular explanations and interventions. However, Concept Splitting is limited to a single additional level of granularity: it can reveal sub-concepts of a provided concept, but it cannot expose deeper hierarchical structure among those sub-concepts. In this work, we introduce MLCS to overcome this limitation.

HiCEMs \citep{hicem} were proposed as an inherently interpretable model for representing hierarchical concept structure, using Concept Splitting to identify parent-child groups. In a HiCEM, from a latent code $\mathbf{h}$, we learn two embeddings per top-level concept. The positive embedding, $\mathbf{\hat{c}}_i^{+\prime}$, represents concept $c_i$’s active state, and the negative embedding, $\mathbf{\hat{c}}_i^{-\prime}$, represents its inactive
state. These embeddings are then passed through sub-concepts modules, which produce new embeddings ($\mathbf{\hat{c}}_i^{+}$ and $\mathbf{\hat{c}}_i^{-}$) that include information about sub-concepts. The sub-concepts modules also output the most likely sub-concept probabilities, which are used to calculate top-level concept probabilities. These probabilities are used to construct an embedding for each concept via a weighted mixture of positive and negative embeddings.

HiCEMs support concept interventions: a user can correct concept predictions at test time, and the downstream task prediction is recomputed using the corrected concepts, enabling targeted human-in-the-loop control. Importantly, the original HiCEM formulation focuses on shallow hierarchies with only two levels (concepts and sub-concepts). This leaves open the question of how to represent and learn deeper, multi-level hierarchies, which we address with Deep-HiCEM, a generalisation that supports hierarchical concept structure beyond a single parent-child layer.

\section{Multi-Level Concept Splitting}
\label{section:mlcs}

Concept Splitting \citep{hicem} discovers fine-grained concepts by training Sparse AutoEncoders (SAEs, \citet{monosemanticity}) on a pretrained CEM’s \citep{cem} concept embeddings and turning sparse features into new concept labels. However, it is limited to a single additional level of granularity: it can reveal sub-concepts of a provided concept, but it cannot expose deeper hierarchical structure among those sub-concepts.

We address this limitation with MLCS, which replaces the single-level SAE with a Hierarchical Sparse AutoEncoder (HiSAE) that learns structured sparse features at multiple levels. This enables the discovery of both sub-concepts and sub-sub-concepts from the same embedding space, allowing us to construct deeper concept hierarchies from only top-level concept supervision.

The use of a HiSAE is the primary difference between MLCS and Concept Splitting, so we focus here on the HiSAE architecture. The HiSAE is designed to learn sparse structure at two or more levels simultaneously: a top level captures candidate sub-concepts (e.g., apple), and a lower level, conditioned on each discovered sub-concept, captures finer-grained refinements (e.g., red apple).

\subsection{Hierarchical Sparse Autoencoder}

The HiSAE, inspired by \citet{muchane}, consists of (i) a top-level encoder that maps an input embedding to a dictionary of size $K$, (ii) a set of sub-encoders, one per top-level latent, each mapping the input to a sub-dictionary of size $K_s$, and (iii) corresponding linear decoders for both levels.

Given an input embedding $\mathbf{e}$, the top-level encoder produces activations over the $K$ top-level latents. A top-$k$ sparsification step \citep{batchtopk} retains only the $k$ largest activations.

For each of the $k$ \textit{active} top-level latents $\ell$, the corresponding sub-encoder processes the input embedding $\mathbf{e}$ to produce activations over its sub-dictionary. A second top-$k_s$ operation selects $k_s$ active sub-latents. The reconstruction is formed by summing the contributions of the active top-level and sub-level decoders, and the model is trained with a standard mean-squared error reconstruction loss.\looseness-1

Crucially, the sub-level is gated by the top-level: sub-latents only contribute when their parent latent is active. This ties sub-sub-concepts to their parent sub-concepts.

\subsection{Depth of the Hierarchy}
In principle, the HiSAE architecture can be extended recursively, yielding arbitrarily deep hierarchies by attaching further sparse autoencoding stages to sub-latents. MLCS therefore provides a general mechanism for hierarchical concept discovery from embedding spaces.

In this work, however, we restrict ourselves to \textit{two discovered levels}: sub-concepts and sub-sub-concepts, to make evaluation more straightforward. Even with this restriction, MLCS allows models to represent richer conceptual structures than single-level Concept Splitting while still requiring supervision only at the top level.

As with Concept Splitting~\citep{hicem}, we can interpret discovered concepts using prototypes, which provide training examples that strongly activate the concept. This approach enables experts to assign potential semantics to discovered concepts.

\section{Deep Hierarchical CEMs}
\label{section:deep-hicems}

\begin{figure}[!tbp]
  \centering
  \includegraphics[width=\textwidth]{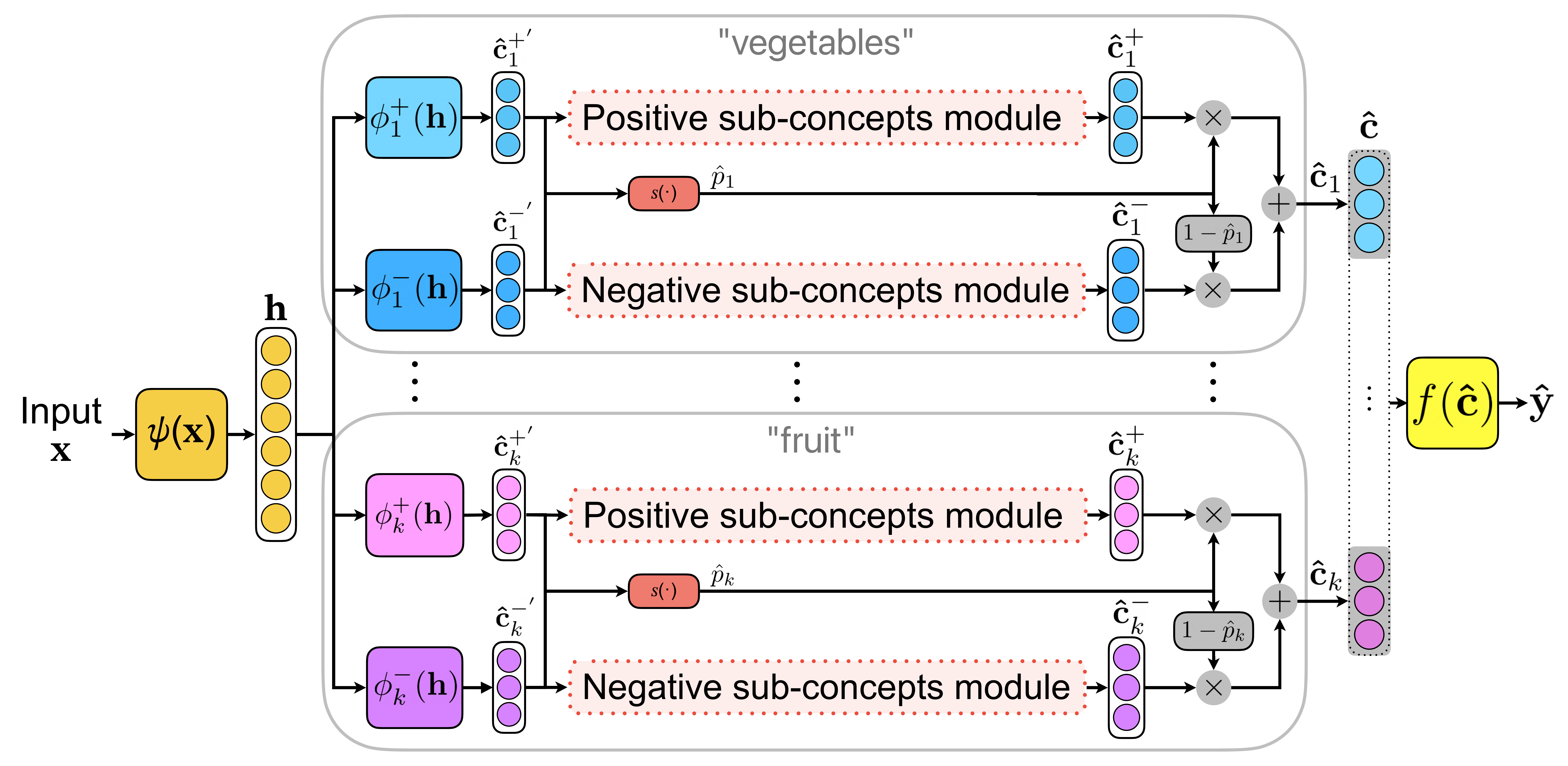}
  \caption{\textbf{Deep-HiCEM:} as in a HiCEM, from a latent code $\mathbf{h}$, we learn two embeddings per concept ($\mathbf{\hat{c}_i^{+\prime}}$ and $\mathbf{\hat{c}_i^{-\prime}}$), which are then passed through sub-concepts modules, which produce new embeddings ($\mathbf{\hat{c}_i^{+}}$ and $\mathbf{\hat{c}_i^{-}}$) that include information about sub-concepts and their descendants.}
  \label{hicem}
\end{figure}

We introduce the Deep-HiCEM architecture (Figure~\ref{hicem}), which extends the HiCEM architecture~\citep{hicem} to support deeper concept hierarchies, like those produced by MLCS. We focus on the differences from HiCEMs; importantly, we do not discuss training Deep-HiCEMs in detail, as the process is the same as in HiCEMs.

\subsection{Concept Structure}

Deep-HiCEMs represent concepts that have been organised into a tree. Each node in the tree represents a concept and can have both \textit{positive} and \textit{negative} sub-concepts. Positive sub-concepts can only be present in an example if their parent concept is present, and negative sub-concepts can only be present if their parent concept is not. The concept trees may be arbitrarily deep.

\subsection{Architecture} \label{hicemarch}

As in HiCEMs, for each top-level concept, a Deep-HiCEM learns a mixture of two embeddings with semantics representing the concept's activity. Each top-level concept $c_i$ is represented with the embeddings $\mathbf{\hat{c}}_i^{+}, \mathbf{\hat{c}}_i^{-} \in \mathbb{R}^m$. Here, $\mathbf{\hat{c}}_i^{+}$ represents $c_i$'s active state, and $\mathbf{\hat{c}}_i^{-}$ represents its inactive state. We also want $\mathbf{\hat{c}}_i^{+}$ and $\mathbf{\hat{c}}_i^{-}$ to contain information about $c_i$'s positive and negative sub-concepts (and their descendants), respectively. To achieve this, a backbone network $\psi(\mathbf{x})$ produces a latent representation $\mathbf{h} \in \mathbb{R}^{n_\text{hidden}}$, which is the input to top-level embedding generators $\phi_i^{+}$ and $\phi_i^{-}$. These top-level embedding generators produce intermediate embeddings $\mathbf{\hat{c}}_i^{+\prime}=\phi_i^+(\mathbf{h}), \mathbf{\hat{c}}_i^{-\prime}=\phi_i^-(\mathbf{h}) \in \mathbb{R}^m$. As in HiCEMs, we implement the top-level embedding generators as single fully connected layers.\looseness-1

To produce final concept embeddings that contain information about sub-concepts, the embeddings $\mathbf{\hat{c}}_i^{+\prime}$ and $\mathbf{\hat{c}}_i^{-\prime}$ are passed through a positive and a negative \textit{sub-concepts module}, respectively. The positive sub-concepts module, which we describe in further detail below, is responsible for learning the positive sub-concepts of $c_i$, and their descendants. It outputs the positive concept embedding for $c_i$, $\mathbf{\hat{c}}_i^{+}$. Similarly, the negative sub-concepts module outputs the negative concept embedding for $c_i$, $\mathbf{\hat{c}}_i^{-}$. If concept $c_i$ has no positive sub-concepts, then we take $\mathbf{\hat{c}}_i^+=\mathbf{\hat{c}}_i^{+\prime}$. We proceed analogously in the absence of negative sub-concepts. The probability of concept $c_i$ is calculated as $\hat{p}_i=s([\mathbf{\hat{c}}_i^{+\prime}, \mathbf{\hat{c}}_i^{-\prime}]^T)$, where $s$ is a shared scoring function that calculates concept probabilities from concept embeddings. As in HiCEMs, the final concept embedding $\mathbf{\hat{c}}_i$ for $c_i$ is calculated as a weighted mixture of $\mathbf{\hat{c}}_i^+$ and $\mathbf{\hat{c}}_i^-$: $\mathbf{\hat{c}}_i = \hat{p}_i\mathbf{\hat{c}}_i^+ + (1 - \hat{p}_i)\mathbf{\hat{c}}_i^-$. Downstream task predictions are calculated as in HiCEMs.\looseness-1

Like HiCEMs, Deep-HiCEMs support concept interventions, such that when a sub-concept is intervened on, its parent may also be updated (e.g., if a human expert informs the model that a positive sub-concept is present, this implies that its parent is also present).

\subsection{Sub-concepts Modules}

We describe a positive sub-concepts module, but negative sub-concepts modules operate in exactly the same way. Sub-concept embeddings are produced in the same way as top-level concept embeddings (Figure~\ref{hicem} in Section~\ref{hicemarch}), with nested sub-concepts modules if the sub-concept has sub-concepts itself. However, instead of operating on the latent code $\mathbf{h}$, sub-concept embedding generators take as input the preliminary embedding of the parent concept. The concatenated mixed embeddings for all the sub-concepts are passed through an embedding compressor (implemented as a single fully connected layer) to produce a fixed-length positive embedding for the parent concept, $\mathbf{\hat{c}}_i^+$.\looseness-1

\section{Experiments}
\label{section:experiments}
We evaluate MLCS and Deep-HiCEMs by exploring the following research questions:
\begin{itemize}[topsep=0pt, leftmargin=25pt, itemsep=0pt]
    \item[{\bf RQ1}] Does MLCS discover interpretable concept hierarchies?

    \item[{\bf RQ2}] How do Deep-HiCEMs' task accuracies compare to those of standard HiCEMs?

    \item[{\bf RQ3}] Can a Deep-HiCEM's task accuracy be improved by intervening on discovered concepts?
\end{itemize}

\subsection{PseudoKitchens-2}
\label{section:pseudokitchens}

To evaluate the ability of MLCS and Deep-HiCEMs to discover and represent two levels of hierarchy (sub-concepts and sub-sub-concepts), we adapt the PseudoKitchens dataset introduced by \citet{hicem} so that both sub-concepts and sub-sub-concepts can be discovered. The discoverable sub-concepts are ingredients (e.g., apple), and the sub-sub-concepts correspond to different variants of that ingredient (e.g., red apple). Full details are given in Appendix~\ref{appendix:PseudoKitchens-2}.

\subsection{Setup}

\paragraph{Datasets} We evaluate our method across five datasets: MNIST-ADD (\citet{hicem}, based on \citet{mnist}), the SHAPES dataset \citep{hicem}, Caltech-UCSD Birds-200-2011 (CUB) \citep{cub}, Animals with Attributes 2 (AwA2) \citep{awa}, and our PseudoKitchens-2 dataset. Full details on PseudoKitchens-2 are included in Appendix~\ref{appendix:PseudoKitchens-2}, and details for the other datasets are given in \citep{hicem}.

\paragraph{Metrics}
Following \citet{hicem}, we first run MLCS on the provided concepts using an initial CEM, and then train a Deep-HiCEM with both the provided top-level concepts and the concepts discovered by MLCS. We evaluate discovered concepts and perform interventions by automatically aligning discovered concepts with a human-interpretable ``left-out'' concept from a predefined ``concept bank''. This bank contains anticipated concepts that are excluded during the initial CEM training (e.g., ``the first digit is 5'' in MNIST-ADD), and each bank concept is associated in advance with the parent concept whose sub-concept it may represent.

To match discovered concepts to the bank, we compute ROC-AUC scores between the discovered concept labels and their potential parent-concept-associated matches in the bank. As in \citet{hicem}, Each concept in the bank is assigned to the sub-concept with the highest ROC-AUC score, as long as that score is greater than 0.7. Discovered concepts without a bank match above this threshold are excluded from evaluation.

For PseudoKitchens-2, the bank contains both sub-concepts (e.g., ``apple'') and sub-sub-concepts (e.g., ``red apple''). When matching discovered sub-concepts to the concept bank, we use the average ROC-AUC of the sub-concept and any associated sub-sub-concepts. For all other datasets, we match and evaluate only sub-concepts, as ground truth sub-sub-concept labels are not available.

To answer \textbf{RQ1}, we report the average discovered concept ROC-AUC of the Deep-HiCEM. For \textbf{RQ2}, we compare task accuracy and provided-concept ROC-AUC between a standard HiCEM and a Deep-HiCEM. For \textbf{RQ3}, we measure the change in the task accuracy of Deep-HiCEMs as concepts are intervened on. All metrics are computed on the test sets using three random seeds, and we report means and standard deviations. Concept accuracy is always summarised using mean ROC-AUC to avoid misleading results from majority-class predictors. For details on model architectures, training and hyperparameters, see Appendix~\ref{appendix:models_training}.

\paragraph{Baselines} We compare our Deep-HiCEMs against standard HiCEMs \citep{hicem}. We also report results for the baselines considered by \citet{hicem}: black-box models, CEMs \citep{cem}, CBMs \citep{cbm}, label-free CBMs (\textit{LF-CBMs}, \citep{labelfree}), Post-hoc CBMs (\textit{PCBMs}, \citep{posthoc}), and PCBMs with residual connections (\textit{PCBM-hs}). For the PseudoKitchens-2 dataset, we implement the baselines following the protocol of \citet{hicem}; for the other datasets, we use the results reported in their work.

\subsection{Results}

\begin{table}[!tbp]
    \caption{Mean ROC-AUC for discovered concepts. LF-CBMs were unable to discover concepts on the MNIST-ADD and PseudoKitchens-2 datasets. Concepts discovered with MLCS are predicted accurately.}
    \label{discoveredaccuracies}
    \centering
    \resizebox{\textwidth}{!}{
    \begin{tabular}{llllll}
        \toprule
           & MNIST-ADD & SHAPES & CUB & AwA2 & PseudoKitchens-2 \\
        \midrule
        LF-CBM & -- & $0.75_{\pm 0.00}$ & $0.77_{\pm 0.00}$ & $0.78_{\pm 0.00}$ & -- \\
        HiCEM + Concept Splitting & $0.93_{\pm 0.01}$ & $\mathbf{0.93_{\pm 0.01}}$ & $\mathbf{0.85_{\pm 0.01}}$ & $\mathbf{0.88_{\pm 0.01}}$ & $\mathbf{0.80_{\pm 0.00}}$ \\
        Deep-HiCEM + MLCS (ours) & $\mathbf{0.94_{\pm 0.01}}$ & $\mathbf{0.93_{\pm 0.01}}$ & $0.84_{\pm 0.01}$ & $0.85_{\pm 0.01}$ & $0.79_{\pm 0.01}$\\
        \bottomrule
    \end{tabular}
    }
    \vspace*{-1mm}
\end{table}

\paragraph{Discovered concepts are human-interpretable and can be predicted accurately (RQ1, Table~\ref{discoveredaccuracies}).} We report the accuracy of the discovered concept predictions made by our models using the ground truth labels of the corresponding human-interpretable concept bank concepts on the test datasets. Table~\ref{discoveredaccuracies} shows that the mean discovered concept \mbox{ROC-AUCs} of Deep-HiCEMs are consistently high, always within a few percentage points of those of HiCEMs. This indicates that the labels produced by MLCS align Deep-HiCEM concept activations with human-interpretable concepts. Matching HiCEMs within a few percentage points is a strong result, given that our method discovers additional levels of hierarchy. On PseudoKitchens-2, the majority of sub-sub-concepts in the bank were successfully matched to discovered sub-sub-concepts on all runs. The mean sub-concept \mbox{ROC-AUC} was $0.80_{\pm 0.00}$, and the mean sub-sub-concept \mbox{ROC-AUC} was $0.79_{\pm 0.01}$. This demonstrates that MLCS and Deep-HiCEMs can discover and represent deeper concept hierarchies while maintaining discovered-concept \mbox{ROC-AUCs} close to those of HiCEMs after Concept Splitting.

\begin{table}[!tbp]
    \caption{Task accuracies. The task accuracy of Deep-HiCEMs is competitive with all our baselines.}
    \label{taskaccuracies}
    \centering
    \resizebox{\textwidth}{!}{
    \begin{tabular}{llllll}
        \toprule
           & MNIST-ADD & SHAPES & CUB & AwA2 & PseudoKitchens-2 \\
        \midrule
        Black box (not interpretable) & $0.94_{\pm 0.00}$ & $0.89_{\pm 0.00}$ & $0.80_{\pm 0.00}$ & $0.98_{\pm 0.00}$ & $0.63_{\pm 0.02}$ \\
        \midrule
        LF-CBM & -- & $0.59_{\pm 0.01}$ & $\mathbf{0.80_{\pm 0.00}}$ & $0.94_{\pm 0.00}$ & -- \\
        PCBM & $0.16_{\pm 0.03}$ & $0.54_{\pm 0.01}$ & $0.65_{\pm 0.01}$ & $0.95_{\pm 0.00}$ & $0.12_{\pm 0.00}$ \\
        PCBM-h & $0.53_{\pm 0.01}$ & $0.73_{\pm 0.00}$ & $0.73_{\pm 0.00}$ & $0.96_{\pm 0.00}$ & $0.48_{\pm 0.00}$ \\
        CBM & $0.23_{\pm 0.01}$ & $0.78_{\pm 0.01}$ & $0.65_{\pm 0.00}$ & $0.97_{\pm 0.00}$ & $\mathbf{0.60_{\pm 0.01}}$ \\
        CEM & $\mathbf{0.92_{\pm 0.01}}$ & $\mathbf{0.89_{\pm 0.00}}$ & $0.76_{\pm 0.01}$ & $\mathbf{0.98_{\pm 0.00}}$ & $0.59_{\pm 0.01}$ \\
        HiCEM + Concept Splitting & $\mathbf{0.92_{\pm 0.00}}$ & $0.87_{\pm 0.02}$ & $0.74_{\pm 0.01}$ & $\mathbf{0.98_{\pm 0.00}}$ & $0.57_{\pm 0.03}$ \\
        Deep-HiCEM + MLCS (ours) & $\mathbf{0.92_{\pm 0.00}}$ & $0.87_{\pm 0.01}$ & $0.73_{\pm 0.01}$ & $0.97_{\pm 0.00}$ & $0.58_{\pm 0.01}$ \\
        \bottomrule
    \end{tabular}
    }
    \vspace*{-1mm}
\end{table}

\begin{table}[!tbp]
    \centering
    \caption{\label{providedaccuracies}Mean ROC-AUCs for provided concepts. Deep-HiCEMs are able to predict provided concepts just as well as HiCEMs.}
    \resizebox{\textwidth}{!}{%
    \begin{tabular}{llllll}
        \toprule
           & MNIST-ADD & SHAPES & CUB & AwA2 & PseudoKitchens-2 \\
        \midrule
        CBM & $\mathbf{0.99_{\pm 0.00}}$ & $\mathbf{1.00_{\pm 0.00}}$ & $0.89_{\pm 0.00}$ & $\mathbf{1.00_{\pm 0.00}}$ & $0.90_{\pm 0.00}$ \\
        CEM & $\mathbf{0.99_{\pm 0.00}}$ & $\mathbf{1.00_{\pm 0.00}}$ & $\mathbf{0.95_{\pm 0.00}}$ & $\mathbf{1.00_{\pm 0.00}}$ & $\mathbf{0.91_{\pm 0.00}}$ \\
        HiCEM + Concept Splitting & $\mathbf{0.99_{\pm 0.00}}$ & $\mathbf{1.00_{\pm 0.00}}$ & $0.93_{\pm 0.01}$ & $\mathbf{1.00_{\pm 0.00}}$ & $\mathbf{0.91_{\pm 0.00}}$ \\
        Deep-HiCEM + MLCS (ours) & $\mathbf{0.99_{\pm 0.00}}$ & $\mathbf{1.00_{\pm 0.00}}$ & $0.94_{\pm 0.00}$ & $\mathbf{1.00_{\pm 0.00}}$ & $\mathbf{0.91_{\pm 0.00}}$ \\
        \bottomrule
    \end{tabular}%
    }
    \vspace*{-1mm}
\end{table}

\paragraph{Deep-HiCEMs have high task and provided concept accuracies (RQ2, Tables~\ref{taskaccuracies}~and~\ref{providedaccuracies}).} We measure the task and provided (top-level) concept accuracies of Deep-HiCEMs and our baselines. The results are in Tables~\ref{taskaccuracies}~and~\ref{providedaccuracies}. Deep-HiCEMs achieve high task and provided concept accuracies compared to the baselines. In particular, the task and provided concept accuracies of Deep-HiCEMs are never more than 1\% below that of standard HiCEMs, so the discovery of deeper hierarchies does not lead to a reduction in task or provided concept accuracy.

\begin{figure}[t!]
    \centering
    \begin{tikzpicture}
    \pgfplotsset{scaled y ticks=false}
    \tikzstyle{every node}=[font=\small]
    \pgfplotsset{footnotesize,samples=10}
    \begin{groupplot}[group style = {group size = 5 by 1, horizontal sep = 25pt}, width = 3.35cm, height = 3.5cm]
        \nextgroupplot[ title = {MNIST-ADD}, xlabel = {Number intervened}, ylabel = {Task accuracy}, legend style = { column sep = 10pt, legend columns = 2, legend to name = grouplegend2,}]
            \addplot+[green, mark options={fill=green}] table[x=n_intervened,y=hicem] {data/mnist_discovered_interventions.dat}; \addlegendentry{HiCEM + Concept Splitting}
            \addplot [name path=upper100,draw=none, forget plot] table[x=n_intervened,y expr=\thisrow{hicem}+\thisrow{hicem_err}] {data/mnist_discovered_interventions.dat};
            \addplot [name path=lower100,draw=none, forget plot] table[x=n_intervened,y expr=\thisrow{hicem}-\thisrow{hicem_err}] {data/mnist_discovered_interventions.dat};
            \addplot [fill=green, fill opacity=0.2, forget plot] fill between[of=upper100 and lower100];

            \addplot+[orange, mark options={fill=orange}] table[x=n_intervened,y=deep_hicem] {data/mnist_discovered_interventions_deep.dat}; \addlegendentry{Deep-HiCEM + MLCS}
            \addplot [name path=upper101,draw=none, forget plot] table[x=n_intervened,y expr=\thisrow{deep_hicem}+\thisrow{deep_hicem_err}] {data/mnist_discovered_interventions_deep.dat};
            \addplot [name path=lower101,draw=none, forget plot] table[x=n_intervened,y expr=\thisrow{deep_hicem}-\thisrow{deep_hicem_err}] {data/mnist_discovered_interventions_deep.dat};
            \addplot [fill=orange, fill opacity=0.2, forget plot] fill between[of=upper101 and lower101];

        \nextgroupplot[title = {SHAPES}, xlabel = {Number intervened}, xtick={0, 5, 10}]
            \addplot+[green, mark options={fill=green}] table[x=n_intervened,y=hicem] {data/shapes_discovered_interventions.dat};
            \addplot [name path=upper103,draw=none, forget plot] table[x=n_intervened,y expr=\thisrow{hicem}+\thisrow{hicem_err}] {data/shapes_discovered_interventions.dat};
            \addplot [name path=lower103,draw=none, forget plot] table[x=n_intervened,y expr=\thisrow{hicem}-\thisrow{hicem_err}] {data/shapes_discovered_interventions.dat};
            \addplot [fill=green, fill opacity=0.2, forget plot] fill between[of=upper103 and lower103];

            \addplot+[orange, mark options={fill=orange}] table[x=n_intervened,y=deep_hicem] {data/shapes_discovered_interventions_deep.dat};
            \addplot [name path=upper104,draw=none, forget plot] table[x=n_intervened,y expr=\thisrow{deep_hicem}+\thisrow{deep_hicem_err}] {data/shapes_discovered_interventions_deep.dat};
            \addplot [name path=lower104,draw=none, forget plot] table[x=n_intervened,y expr=\thisrow{deep_hicem}-\thisrow{deep_hicem_err}] {data/shapes_discovered_interventions_deep.dat};
            \addplot [fill=orange, fill opacity=0.2, forget plot] fill between[of=upper104 and lower104];

        \nextgroupplot[title = {CUB}, xlabel = {Number intervened}, xtick={0, 75, 150}, each nth point=10, filter discard warning=false, unbounded coords=discard] 
            \addplot+[green, mark options={fill=green}] table[x=n_intervened,y=hicem] {data/cub_discovered_interventions.dat};
            \addplot [name path=upper106,draw=none, forget plot] table[x=n_intervened,y expr=\thisrow{hicem}+\thisrow{hicem_err}] {data/cub_discovered_interventions.dat};
            \addplot [name path=lower106,draw=none, forget plot] table[x=n_intervened,y expr=\thisrow{hicem}-\thisrow{hicem_err}] {data/cub_discovered_interventions.dat};
            \addplot [fill=green, fill opacity=0.2, forget plot] fill between[of=upper106 and lower106];

            \addplot+[orange, mark options={fill=orange}] table[x=n_intervened,y=deep_hicem] {data/cub_discovered_interventions_deep.dat};
            \addplot [name path=upper107,draw=none, forget plot] table[x=n_intervened,y expr=\thisrow{deep_hicem}+\thisrow{deep_hicem_err}] {data/cub_discovered_interventions_deep.dat};
            \addplot [name path=lower107,draw=none, forget plot] table[x=n_intervened,y expr=\thisrow{deep_hicem}-\thisrow{deep_hicem_err}] {data/cub_discovered_interventions_deep.dat};
            \addplot [fill=orange, fill opacity=0.2, forget plot] fill between[of=upper107 and lower107];

        \nextgroupplot[title = {AwA2}, xlabel = {Number intervened}, ] 
            \addplot+[green, mark options={fill=green}] table[x=n_intervened,y=hicem] {data/awa_discovered_interventions.dat};
            \addplot [name path=upper109,draw=none, forget plot] table[x=n_intervened,y expr=\thisrow{hicem}+\thisrow{hicem_err}] {data/awa_discovered_interventions.dat};
            \addplot [name path=lower109,draw=none, forget plot] table[x=n_intervened,y expr=\thisrow{hicem}-\thisrow{hicem_err}] {data/awa_discovered_interventions.dat};
            \addplot [fill=green, fill opacity=0.2, forget plot] fill between[of=upper109 and lower109];

            \addplot+[orange, mark options={fill=orange}] table[x=n_intervened,y=deep_hicem] {data/awa_discovered_interventions_deep.dat};
            \addplot [name path=upper110,draw=none, forget plot] table[x=n_intervened,y expr=\thisrow{deep_hicem}+\thisrow{deep_hicem_err}] {data/awa_discovered_interventions_deep.dat};
            \addplot [name path=lower110,draw=none, forget plot] table[x=n_intervened,y expr=\thisrow{deep_hicem}-\thisrow{deep_hicem_err}] {data/awa_discovered_interventions_deep.dat};
            \addplot [fill=orange, fill opacity=0.2, forget plot] fill between[of=upper110 and lower110];

        \nextgroupplot[title = {PseudoKitchens-2}, xlabel = {Number intervened}] 
            \addplot+[green, mark options={fill=green}] table[x=n_intervened,y=hicem] {data/kitchens_discovered_interventions.dat};
            \addplot [name path=upper9109,draw=none, forget plot] table[x=n_intervened,y expr=\thisrow{hicem}+\thisrow{hicem_err}] {data/kitchens_discovered_interventions.dat};
            \addplot [name path=lower9109,draw=none, forget plot] table[x=n_intervened,y expr=\thisrow{hicem}-\thisrow{hicem_err}] {data/kitchens_discovered_interventions.dat};
            \addplot [fill=green, fill opacity=0.2, forget plot] fill between[of=upper9109 and lower9109];

            \addplot+[orange, mark options={fill=orange}] table[x=n_intervened,y=deep_hicem] {data/kitchens_discovered_interventions_deep.dat};
            \addplot [name path=upper9110,draw=none, forget plot] table[x=n_intervened,y expr=\thisrow{deep_hicem}+\thisrow{deep_hicem_err}] {data/kitchens_discovered_interventions_deep.dat};
            \addplot [name path=lower9110,draw=none, forget plot] table[x=n_intervened,y expr=\thisrow{deep_hicem}-\thisrow{deep_hicem_err}] {data/kitchens_discovered_interventions_deep.dat};
            \addplot [fill=orange, fill opacity=0.2, forget plot] fill between[of=upper9110 and lower9110];

    \end{groupplot}
    \node at ($(group c2r1) + (2.6,-2.3cm)$) {\ref{grouplegend2}}; 
    \end{tikzpicture}
    \caption{Task accuracy as discovered concepts are intervened on. Intervening on discovered concepts improves task accuracy, with a few exceptions that would benefit from further investigation.}
    \label{discoveredinterventions}
\end{figure}
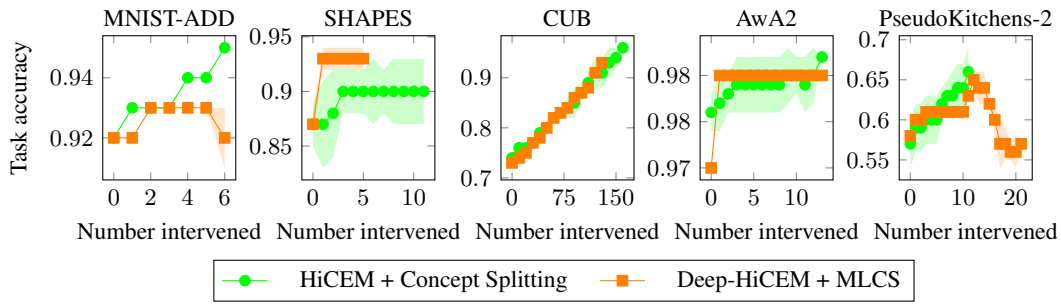

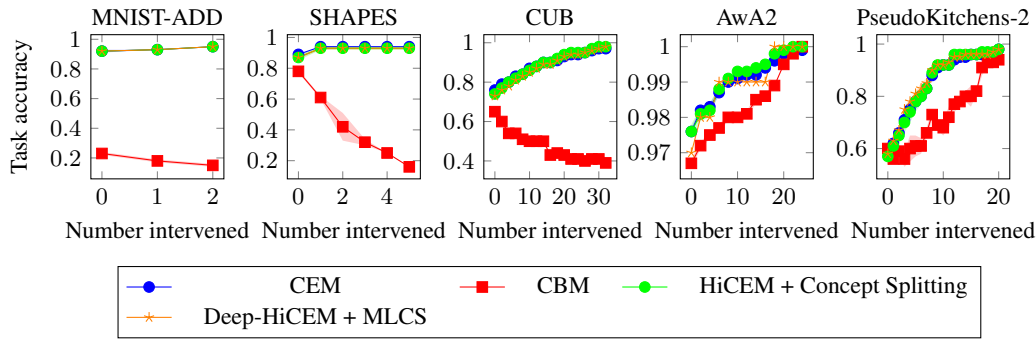
\begin{figure}[t!]
    \centering
    \begin{tikzpicture}
    \tikzstyle{every node}=[font=\small]
    \pgfplotsset{footnotesize,samples=10}
    \pgfplotsset{scaled y ticks=false}
    \begin{groupplot}[group style = {group size = 5 by 1, horizontal sep = 24pt}, width = 3.35cm, height = 3.5cm]
        \nextgroupplot[ title = {MNIST-ADD}, xlabel = {Number intervened}, ylabel = {Task accuracy},
            legend style = { column sep = 10pt, legend columns = 3, legend to name = grouplegend1,},xtick={0, 1, 2}]

            \addplot+[blue, mark options={fill=blue}] table[x=n_intervened,y=cem] {data/mnist_provided_interventions.dat}; \addlegendentry{CEM}
            \addplot [name path=upper102,draw=none, forget plot] table[x=n_intervened,y expr=\thisrow{cem}+\thisrow{cem_err}] {data/mnist_provided_interventions.dat};
            \addplot [name path=lower102,draw=none, forget plot] table[x=n_intervened,y expr=\thisrow{cem}-\thisrow{cem_err}] {data/mnist_provided_interventions.dat};
            \addplot [fill=blue, fill opacity=0.2, forget plot] fill between[of=upper102 and lower102];

            \addplot+[red, mark options={fill=red}] table[x=n_intervened,y=cbm] {data/mnist_provided_interventions.dat}; \addlegendentry{CBM}
            \addplot [name path=upper103,draw=none, forget plot] table[x=n_intervened,y expr=\thisrow{cbm}+\thisrow{cbm_err}] {data/mnist_provided_interventions.dat};
            \addplot [name path=lower103,draw=none, forget plot] table[x=n_intervened,y expr=\thisrow{cbm}-\thisrow{cbm_err}] {data/mnist_provided_interventions.dat};
            \addplot [fill=red, fill opacity=0.2, forget plot] fill between[of=upper103 and lower103];

            \addplot+[green, mark options={fill=green}] table[x=n_intervened,y=hicem] {data/mnist_provided_interventions.dat}; \addlegendentry{HiCEM + Concept Splitting}
            \addplot [name path=upper100,draw=none, forget plot] table[x=n_intervened,y expr=\thisrow{hicem}+\thisrow{hicem_err}] {data/mnist_provided_interventions.dat};
            \addplot [name path=lower100,draw=none, forget plot] table[x=n_intervened,y expr=\thisrow{hicem}-\thisrow{hicem_err}] {data/mnist_provided_interventions.dat};
            \addplot [fill=green, fill opacity=0.2, forget plot] fill between[of=upper100 and lower100];

            \addplot+[orange, mark options={fill=orange}] table[x=n_intervened,y=deep_hicem] {data/mnist_provided_interventions_deep.dat}; \addlegendentry{Deep-HiCEM + MLCS}
            \addplot [name path=upper101,draw=none, forget plot] table[x=n_intervened,y expr=\thisrow{deep_hicem}+\thisrow{deep_hicem_err}] {data/mnist_provided_interventions_deep.dat};
            \addplot [name path=lower101,draw=none, forget plot] table[x=n_intervened,y expr=\thisrow{deep_hicem}-\thisrow{deep_hicem_err}] {data/mnist_provided_interventions_deep.dat};
            \addplot [fill=orange, fill opacity=0.2, forget plot] fill between[of=upper101 and lower101];

        \nextgroupplot[title = {SHAPES}, xlabel = {Number intervened}]

            \addplot+[blue, mark options={fill=blue}] table[x=n_intervened,y=cem] {data/shapes_provided_interventions.dat};
            \addplot [name path=upper105,draw=none, forget plot] table[x=n_intervened,y expr=\thisrow{cem}+\thisrow{cem_err}] {data/shapes_provided_interventions.dat};
            \addplot [name path=lower105,draw=none, forget plot] table[x=n_intervened,y expr=\thisrow{cem}-\thisrow{cem_err}] {data/shapes_provided_interventions.dat};
            \addplot [fill=blue, fill opacity=0.2, forget plot] fill between[of=upper105 and lower105];

            \addplot+[red, mark options={fill=red}] table[x=n_intervened,y=cbm] {data/shapes_provided_interventions.dat};
            \addplot [name path=upper106,draw=none, forget plot] table[x=n_intervened,y expr=\thisrow{cbm}+\thisrow{cbm_err}] {data/shapes_provided_interventions.dat};
            \addplot [name path=lower106,draw=none, forget plot] table[x=n_intervened,y expr=\thisrow{cbm}-\thisrow{cbm_err}] {data/shapes_provided_interventions.dat};
            \addplot [fill=red, fill opacity=0.2, forget plot] fill between[of=upper106 and lower106];

            \addplot+[green, mark options={fill=green}] table[x=n_intervened,y=hicem] {data/shapes_provided_interventions.dat};
            \addplot [name path=upper104,draw=none, forget plot] table[x=n_intervened,y expr=\thisrow{hicem}+\thisrow{hicem_err}] {data/shapes_provided_interventions.dat};
            \addplot [name path=lower104,draw=none, forget plot] table[x=n_intervened,y expr=\thisrow{hicem}-\thisrow{hicem_err}] {data/shapes_provided_interventions.dat};
            \addplot [fill=green, fill opacity=0.2, forget plot] fill between[of=upper104 and lower104];

            \addplot+[orange, mark options={fill=orange}] table[x=n_intervened,y=deep_hicem] {data/shapes_provided_interventions_deep.dat};
            \addplot [name path=upper1101,draw=none, forget plot] table[x=n_intervened,y expr=\thisrow{deep_hicem}+\thisrow{deep_hicem_err}] {data/shapes_provided_interventions_deep.dat};
            \addplot [name path=lower1101,draw=none, forget plot] table[x=n_intervened,y expr=\thisrow{deep_hicem}-\thisrow{deep_hicem_err}] {data/shapes_provided_interventions_deep.dat};
            \addplot [fill=orange, fill opacity=0.2, forget plot] fill between[of=upper1101 and lower1101];

        \nextgroupplot[title = {CUB}, xlabel = {Number intervened}, each nth point=2, filter discard warning=false, unbounded coords=discard]

            \addplot+[blue, mark options={fill=blue}] table[x=n_intervened,y=cem] {data/cub_provided_interventions.dat};
            \addplot [name path=upper108,draw=none, forget plot] table[x=n_intervened,y expr=\thisrow{cem}+\thisrow{cem_err}] {data/cub_provided_interventions.dat};
            \addplot [name path=lower108,draw=none, forget plot] table[x=n_intervened,y expr=\thisrow{cem}-\thisrow{cem_err}] {data/cub_provided_interventions.dat};
            \addplot [fill=blue, fill opacity=0.2, forget plot] fill between[of=upper108 and lower108];

            \addplot+[red, mark options={fill=red}] table[x=n_intervened,y=cbm] {data/cub_provided_interventions.dat};
            \addplot [name path=upper109,draw=none, forget plot] table[x=n_intervened,y expr=\thisrow{cbm}+\thisrow{cbm_err}] {data/cub_provided_interventions.dat};
            \addplot [name path=lower109,draw=none, forget plot] table[x=n_intervened,y expr=\thisrow{cbm}-\thisrow{cbm_err}] {data/cub_provided_interventions.dat};
            \addplot [fill=red, fill opacity=0.2, forget plot] fill between[of=upper109 and lower109];

            \addplot+[green, mark options={fill=green}] table[x=n_intervened,y=hicem] {data/cub_provided_interventions.dat};
            \addplot [name path=upper107,draw=none, forget plot] table[x=n_intervened,y expr=\thisrow{hicem}+\thisrow{hicem_err}] {data/cub_provided_interventions.dat};
            \addplot [name path=lower107,draw=none, forget plot] table[x=n_intervened,y expr=\thisrow{hicem}-\thisrow{hicem_err}] {data/cub_provided_interventions.dat};
            \addplot [fill=green, fill opacity=0.2, forget plot] fill between[of=upper107 and lower107];

            \addplot+[orange, mark options={fill=orange}] table[x=n_intervened,y=deep_hicem] {data/cub_provided_interventions_deep.dat};
            \addplot [name path=upper2101,draw=none, forget plot] table[x=n_intervened,y expr=\thisrow{deep_hicem}+\thisrow{deep_hicem_err}] {data/cub_provided_interventions_deep.dat};
            \addplot [name path=lower2101,draw=none, forget plot] table[x=n_intervened,y expr=\thisrow{deep_hicem}-\thisrow{deep_hicem_err}] {data/cub_provided_interventions_deep.dat};
            \addplot [fill=orange, fill opacity=0.2, forget plot] fill between[of=upper2101 and lower2101];

        \nextgroupplot[title = {AwA2}, xlabel = {Number intervened}, each nth point=2, filter discard warning=false, unbounded coords=discard] 

            \addplot+[blue, mark options={fill=blue}] table[x=n_intervened,y=cem] {data/awa_provided_interventions.dat};
            \addplot [name path=upper111,draw=none, forget plot] table[x=n_intervened,y expr=\thisrow{cem}+\thisrow{cem_err}] {data/awa_provided_interventions.dat};
            \addplot [name path=lower111,draw=none, forget plot] table[x=n_intervened,y expr=\thisrow{cem}-\thisrow{cem_err}] {data/awa_provided_interventions.dat};
            \addplot [fill=blue, fill opacity=0.2, forget plot] fill between[of=upper111 and lower111];

            \addplot+[red, mark options={fill=red}] table[x=n_intervened,y=cbm] {data/awa_provided_interventions.dat};
            \addplot [name path=upper112,draw=none, forget plot] table[x=n_intervened,y expr=\thisrow{cbm}+\thisrow{cbm_err}] {data/awa_provided_interventions.dat};
            \addplot [name path=lower112,draw=none, forget plot] table[x=n_intervened,y expr=\thisrow{cbm}-\thisrow{cbm_err}] {data/awa_provided_interventions.dat};
            \addplot [fill=red, fill opacity=0.2, forget plot] fill between[of=upper112 and lower112];

            \addplot+[green, mark options={fill=green}] table[x=n_intervened,y=hicem] {data/awa_provided_interventions.dat};
            \addplot [name path=upper110,draw=none, forget plot] table[x=n_intervened,y expr=\thisrow{hicem}+\thisrow{hicem_err}] {data/awa_provided_interventions.dat};
            \addplot [name path=lower110,draw=none, forget plot] table[x=n_intervened,y expr=\thisrow{hicem}-\thisrow{hicem_err}] {data/awa_provided_interventions.dat};
            \addplot [fill=green, fill opacity=0.2, forget plot] fill between[of=upper110 and lower110];

            \addplot+[orange, mark options={fill=orange}] table[x=n_intervened,y=deep_hicem] {data/awa_provided_interventions_deep.dat};
            \addplot [name path=upper3101,draw=none, forget plot] table[x=n_intervened,y expr=\thisrow{deep_hicem}+\thisrow{deep_hicem_err}] {data/awa_provided_interventions_deep.dat};
            \addplot [name path=lower3101,draw=none, forget plot] table[x=n_intervened,y expr=\thisrow{deep_hicem}-\thisrow{deep_hicem_err}] {data/awa_provided_interventions_deep.dat};
            \addplot [fill=orange, fill opacity=0.2, forget plot] fill between[of=upper3101 and lower3101];

        \nextgroupplot[title = {PseudoKitchens-2}, xlabel = {Number intervened}] 

            \addplot+[blue, mark options={fill=blue}] table[x=n_intervened,y=cem] {data/kitchens_provided_interventions.dat};
            \addplot [name path=upper114,draw=none, forget plot] table[x=n_intervened,y expr=\thisrow{cem}+\thisrow{cem_err}] {data/kitchens_provided_interventions.dat};
            \addplot [name path=lower114,draw=none, forget plot] table[x=n_intervened,y expr=\thisrow{cem}-\thisrow{cem_err}] {data/kitchens_provided_interventions.dat};
            \addplot [fill=blue, fill opacity=0.2, forget plot] fill between[of=upper114 and lower114];

            \addplot+[red, mark options={fill=red}] table[x=n_intervened,y=cbm] {data/kitchens_provided_interventions.dat};
            \addplot [name path=upper115,draw=none, forget plot] table[x=n_intervened,y expr=\thisrow{cbm}+\thisrow{cbm_err}] {data/kitchens_provided_interventions.dat};
            \addplot [name path=lower115,draw=none, forget plot] table[x=n_intervened,y expr=\thisrow{cbm}-\thisrow{cbm_err}] {data/kitchens_provided_interventions.dat};
            \addplot [fill=red, fill opacity=0.2, forget plot] fill between[of=upper115 and lower115];

            \addplot+[green, mark options={fill=green}] table[x=n_intervened,y=hicem] {data/kitchens_provided_interventions.dat};
            \addplot [name path=upper113,draw=none, forget plot] table[x=n_intervened,y expr=\thisrow{hicem}+\thisrow{hicem_err}] {data/kitchens_provided_interventions.dat};
            \addplot [name path=lower113,draw=none, forget plot] table[x=n_intervened,y expr=\thisrow{hicem}-\thisrow{hicem_err}] {data/kitchens_provided_interventions.dat};
            \addplot [fill=green, fill opacity=0.2, forget plot] fill between[of=upper113 and lower113];

            \addplot+[orange, mark options={fill=orange}] table[x=n_intervened,y=deep_hicem] {data/kitchens_provided_interventions_deep.dat};
            \addplot [name path=upper4101,draw=none, forget plot] table[x=n_intervened,y expr=\thisrow{deep_hicem}+\thisrow{deep_hicem_err}] {data/kitchens_provided_interventions_deep.dat};
            \addplot [name path=lower4101,draw=none, forget plot] table[x=n_intervened,y expr=\thisrow{deep_hicem}-\thisrow{deep_hicem_err}] {data/kitchens_provided_interventions_deep.dat};
            \addplot [fill=orange, fill opacity=0.2, forget plot] fill between[of=upper4101 and lower4101];

    \end{groupplot}
    \node at ($(group c2r1) + (2.6,-2.6cm)$) {\ref{grouplegend1}}; 
    \end{tikzpicture}
    \caption{Change in task accuracy as provided concepts are intervened on. Provided concept interventions work just as well in Deep-HiCEMs as they do in HiCEMs.}
    \label{providedinterventions}
\end{figure}

\paragraph{Intervening on concepts discovered by MLCS can enhance task accuracy, with a few exceptions that would benefit from further investigation (RQ3, Figure~\ref{discoveredinterventions})} As shown in Figure~\ref{discoveredinterventions}, intervening on discovered concepts can lead to an increase in task accuracy, although interventions on some discovered concepts have no effect or decrease task accuracy. The decreases in task accuracy caused by discovered concept interventions, particularly in PseudoKitchens-2, would benefit from further investigation, and we suggest this for future work. We
hypothesise that these decreases are due to inaccuracies or biases in the discovered concept labels. We evaluate interventions on the provided top-level concepts in Figure~\ref{providedinterventions} and find that provided concept interventions perform equally well in \mbox{Deep-HiCEMs} and HiCEMs. Overall, these results indicate that many of the concepts uncovered by MLCS are not only interpretable but also actionable, and that leveraging them through interventions can translate into performance gains.

\section{Limitations and Conclusion}
\label{section:discussion}
We introduced MLCS and Deep-HiCEMs, a framework for discovering and modelling deep concept hierarchies from only top-level supervision. Across several datasets, we showed that our approach reliably uncovers human-interpretable concept hierarchies absent during training while maintaining competitive task accuracy. However, interventions on discovered concepts are not consistently beneficial and can sometimes reduce accuracy, motivating further study of when and why this happens. Another limitation of our approach is that SAEs are not guaranteed to discover meaningful concepts. Future work includes scaling to larger, more complex datasets, and extending evaluations to investigate even deeper hierarchies. Despite the limitations mentioned, modelling deeper concept structure is a practical step towards more expressive concept-based interpretability.

\subsubsection*{Acknowledgments}
This work was supported by the Engineering and Physical Sciences Research Council [EP/Y030826/1]. A significant portion of this work was carried out whilst MEZ was at the University of Cambridge, funded by the Gates Cambridge Trust via a Gates Cambridge Scholarship.

\bibliography{iclr2026_conference}
\bibliographystyle{iclr2026_conference}

\newpage
\appendix

\section{PseudoKitchens-2}
\label{appendix:PseudoKitchens-2}

This appendix describes PseudoKitchens-2, an extension of the PseudoKitchens dataset \citep{hicem} designed to enable the discovery and evaluation of two-level concept hierarchies.

PseudoKitchens-2 is generated in the same way as PseudoKitchens \citep{hicem}. The recipes described by \citet{hicem} are altered so that different variations of the same ingredient are distinguished (e.g., some recipes require ``red apples'' while others require ``green apples''). The ingredients that have variations are apple (5 variations), potato (5 variations), and pepper (3 variations).

\subsection{Recipes}
\label{recipes}

We adapted the recipes used by \citet{hicem} that define valid combinations of ingredients for the classification task. Some ingredients are organised into groups as shown in Table~\ref{igroups} (these groups are the same as the ones used by \citet{hicem}). If a recipe contains an ingredient group, a random number of ingredients are selected from that group, unless the group is pasta, in which case only one type of pasta is selected. The recipes used in PseudoKitchens-2 are shown in Table~\ref{tab:recipes}. Where a list of numbers is specified in brackets after an ingredient (e.g., ``Apple (1)''), this refers to the variants of that ingredient that are acceptable. For each instance, a recipe is chosen uniformly at random.

\begin{table}[h]
  \caption{Ingredient groups in PseudoKitchens and PseudoKitchens-2.}
  \label{igroups}
  \centering
  \begin{tabular}{ll}
    \toprule
    Group & Ingredients \\
    \midrule
    Fruit & Banana, Orange, Apple, Pear, Pineapple \\
    Vegetables & Onion, Carrot, Potato, Pepper, Courgette \\
    Pasta & Macaroni, Spaghetti \\
    \bottomrule
  \end{tabular}
\end{table}

\begin{table}[h]
  \caption{Recipes in PseudoKitchens-2.}
  \label{tab:recipes}
  \centering
  \begin{tabular}{ll}
    \toprule
    Recipe & Ingredients \\
    \midrule
    Fruit Salad & Fruit \\
    Vegetable Pasta & Pasta, Onion, Garlic, Oil, Vegetables, Spice, Tin Tomatoes \\
    Risotto & Cheese, Onion, Garlic, Vegetables, Oil, Spice, Rice \\
    Chips & Potato (2, 3, 4), Oil, Flour, Garlic, Spice \\
    Chilli & Mince, Oil, Onion, Garlic, Chilli, Tin Tomatoes, Spice, Rice \\
    Smoothie & Milk, Yoghurt, Fruit \\
    Hot Chocolate & Chocolate, Milk \\
    Banana Bread & Butter, Sugar, Egg, Flour, Banana \\
    Chocolate Fudge Cake & Egg, Sugar, Oil, Flour, Chocolate, Syrup, Milk \\
    Carbonara & Garlic, Meat, Butter, Cheese, Egg, Spaghetti, Spice \\
    Apple Crumble & Apple (3), Sugar, Flour, Butter \\
    Salad & Pepper (2, 3), Apple (1, 2, 4, 5), Potato (1, 5) \\
    \bottomrule
  \end{tabular}
\end{table}

The concepts provided to the initial CEM are the ingredient groups in Table~\ref{igroups} (e.g., ``contains fruit''), as well as all the ingredients that are not part of a group. The concept bank sub-concepts correspond to the ingredients in the ingredient groups (e.g., ``contains apples''), and the sub-sub-concepts correspond to variations of the same ingredient (e.g., ``contains green apples'').

\subsection{Dataset Composition}

Like the original PseudoKitchens, the complete PseudoKitchens-2 dataset comprises 10,000 training images, 1,000 validation images, and 1,000 test images. Each image is rendered at $512 \times 512$ resolution.

\section{Model architectures, training and hyperparameters}
\label{appendix:models_training}

\paragraph{Model architectures} We use the CLIP ViT-L/14 foundation model \citep{clip} as the backbone for all of our models and baselines. We do not fine-tune the foundation model: we just use the representations it outputs. For all the models, we precompute the representations with standard image preprocessing pipelines and do not use any data augmentations.

When training BatchTopK SAEs, like \citet{hicem}, we use the default hyperparameters in the code released by \cite{batchtopk} for all datasets.\footnote{\url{https://github.com/bartbussmann/BatchTopK/blob/main/config.py}} The key distinction of this method is its enforcement of sparsity: it selects the top $n \cdot k$ activations across an entire batch of $n$ samples. This allows the number of active features to vary per sample, targeting an average of $k=32$ active features. The SAEs are trained for 300 epochs with a dictionary size of 12,288 and a learning rate of \num{3e-4}.

For MLCS, we train Hierarchical Sparse Autoencoders (HiSAEs) to discover concept hierarchies. The HiSAEs use a top-level dictionary of size 4096 with top-$k$ sparsification ($k=32$), and for each top-level latent we learn a sub-dictionary of size 512 with top-$k_s$ sparsification ($k_s=16$). HiSAEs are trained with a batch size of 1000 for up to 100 epochs.

All of our CEM and HiCEM concept embeddings have $m=16$ activations. Across all datasets, we always use a single fully connected layer for label predictor $f$.

\paragraph{Training hyperparameters} Our models are trained using the Adam optimisation algorithm \citep{adam} with a learning rate of \num{1e-3}. They are trained for a maximum of 300 epochs, and training is stopped if the validation loss does not improve for 75 epochs. We use a batch size of 256.

In all CEMs, (Deep-)HiCEMs and CBMs, the weight of the concept loss is set to $\lambda=10$. Following \citep{cbm}, in MNIST-ADD, CUB and PseudoKitchens-2 we use a weighted cross-entropy loss for concept prediction to mitigate imbalances in concept labels. In MNIST-ADD, we also use a weighted cross-entropy loss for task prediction to mitigate imbalances in task labels.

When training CEMs and (Deep-)HiCEMs, the RandInt \citep{cem} regularisation strategy is used: at training time, concepts are intervened independently at random, with the probability of an intervention being $p_\text{int}=0.25$. We choose  $p_\text{int}=0.25$ because \citet{cem} find that it enables effective interventions while giving good performance.

\section{Code, licenses, and resources}
\label{appendix:code_licences_resources}

\paragraph{Assets} We used the CLIP foundation models \citep{clip} (\url{https://github.com/openai/CLIP}), whose code is available under the MIT license. To run our experiments, we made use of the CEM \citep{cem} (\url{https://github.com/mateoespinosa/cem}, MIT license), Post-hoc CBM \citep{posthoc} (\url{https://github.com/mertyg/post-hoc-cbm}, MIT license) and Label-free CBM \citep{labelfree} (\url{https://github.com/Trustworthy-ML-Lab/Label-free-CBM}) repositories. We implemented our experiments in Python 3.12 and used open-source libraries such as PyTorch 2.9 \citep{pytorch} (BSD license) and Scikit-learn \citep{sklearn} (BSD license). We have released the code required to recreate our experiments in a MIT-licensed public repository.\footnote{\url{https://github.com/OscarPi/cem-concept-discovery}}

\paragraph{Resources} All of our experiments were run on virtual machines with at least 8 CPU cores, 18GB of RAM, and an NVIDIA GPU (Quadro RTX 8000 or GeForce RTX 4090). We estimate that approximately 100 GPU hours were required to complete our work.

\paragraph{Use of AI} We used Large Language Models (LLMs) as assistants for drafting and improving the clarity and grammar of this manuscript. LLMs were also used to generate boilerplate code. However, all core research ideas, experimental design, and analysis of the results were conducted by the authors.

\end{document}